\documentclass[letterpaper]{article} 
\usepackage{aaai25}  
\usepackage{times}  
\usepackage{helvet}  
\usepackage{courier}  
\usepackage[hyphens]{url}  
\usepackage{graphicx} 
\usepackage{natbib}  
\usepackage{caption} 
\usepackage{algorithm}
\usepackage{algorithmic}

\usepackage{amsmath}
\usepackage{amssymb}
\usepackage{mathrsfs}
\usepackage{dsfont}
\usepackage{pifont}

\usepackage{graphics}
\usepackage{graphicx}
\usepackage{multirow}
\usepackage{multicol}

\usepackage{color,colortbl}
\definecolor{Gray}{gray}{0.9}

\usepackage{newfloat}
\usepackage{listings}
\DeclareCaptionStyle{ruled}{labelfont=normalfont,labelsep=colon,strut=off} 
\floatstyle{ruled}
\newfloat{listing}{tb}{lst}{}
\floatname{listing}{Listing}
\title{SCKD: Semi-Supervised Cross-Modality Knowledge Distillation for 4D Radar Object Detection}
\author{
    Ruoyu Xu\textsuperscript{\rm 1},
    Zhiyu Xiang\textsuperscript{\rm 1,2}\thanks{Corresponding author.},
    Chenwei Zhang\textsuperscript{\rm 1},
    Hanzhi Zhong\textsuperscript{\rm 1},
    Xijun Zhao\textsuperscript{\rm 3},
    Ruina Dang\textsuperscript{\rm 3},    
    Peng Xu\textsuperscript{\rm 1},
    Tianyu Pu\textsuperscript{\rm 1},
    Eryun Liu\textsuperscript{\rm 1}
}
\affiliations{
    \textsuperscript{\rm 1}Zhejiang University, China\\
    \textsuperscript{\rm 2}Zhejiang Provincial Key Laboratory of Multi-Modal Communication Networks and Intelligent Information Processing\\

    \textsuperscript{\rm 3}China North Artificial Intelligence \& Innovation Research Institude\\

    \{xuruoyu, xiangzy, zhangchenwei, zhonghanzhi, xxxupeng, 3190105835, eryunliu\}@zju.edu.cn \\
    \{heejunzhao, ruinadang\}@163.com
}

\usepackage{bibentry}

\begin{document}

\maketitle

\begin{abstract}
  3D object detection is one of the fundamental perception tasks for autonomous vehicles. Fulfilling such a task with a 4D millimeter-wave radar is very attractive since the sensor is able to acquire 3D point clouds similar to Lidar while maintaining robust measurements under adverse weather. However, due to the high sparsity and noise associated with the radar point clouds, the performance of the existing methods is still much lower than expected. In this paper, we propose a novel \textbf{S}emi-supervised \textbf{C}ross-modality \textbf{K}nowledge \textbf{D}istillation (\textbf{SCKD}) method for 4D radar-based 3D object detection. It characterizes the capability of learning the feature from a Lidar-radar-fused teacher network with semi-supervised distillation. We first propose an adaptive fusion module in the teacher network to boost its performance. Then, two feature distillation modules are designed to facilitate the cross-modality knowledge transfer. Finally, a semi-supervised output distillation is proposed to increase the effectiveness and flexibility of the distillation framework. With the same network structure, our radar-only student trained by SCKD boosts the mAP by 10.38\% over the baseline and outperforms the state-of-the-art works on the VoD dataset. The experiment on ZJUODset also shows 5.12\% mAP improvements on the moderate difficulty level over the baseline when extra unlabeled data are available. Code is available at https://github.com/Ruoyu-Xu/SCKD. 
\end{abstract}

%

\section{Introduction}

3D object detection is an essential perception task for autonomous vehicles operating in real traffic scenes. Thanks to the dense point clouds acquired and the development of deep learning technology, Lidar-based 3D object detection methods\cite{yang20203dssd,yan2018second,lang2019pointpillars,yin2021center} have made remarkable progress in recent years. However, constrained by its short wavelength, Lidar performs poorly in adverse weather conditions such as rain and fog\cite{sheeny2021radiate}. In contrast, millimeter-wave radar has gained widespread attention due to its resistance to adverse weather and long measurement distance. Traditional 3D millimeter-wave radar could merely provide two-dimensional point clouds and Doppler velocity, which makes it difficult for 3D object detection. In recent years, with the development of high-resolution 4D millimeter-wave radar, exploring the 4D radar point clouds for scene understanding task becomes very attractive.

\begin{figure}[t]
\centerline{\includegraphics[width=0.93\linewidth]{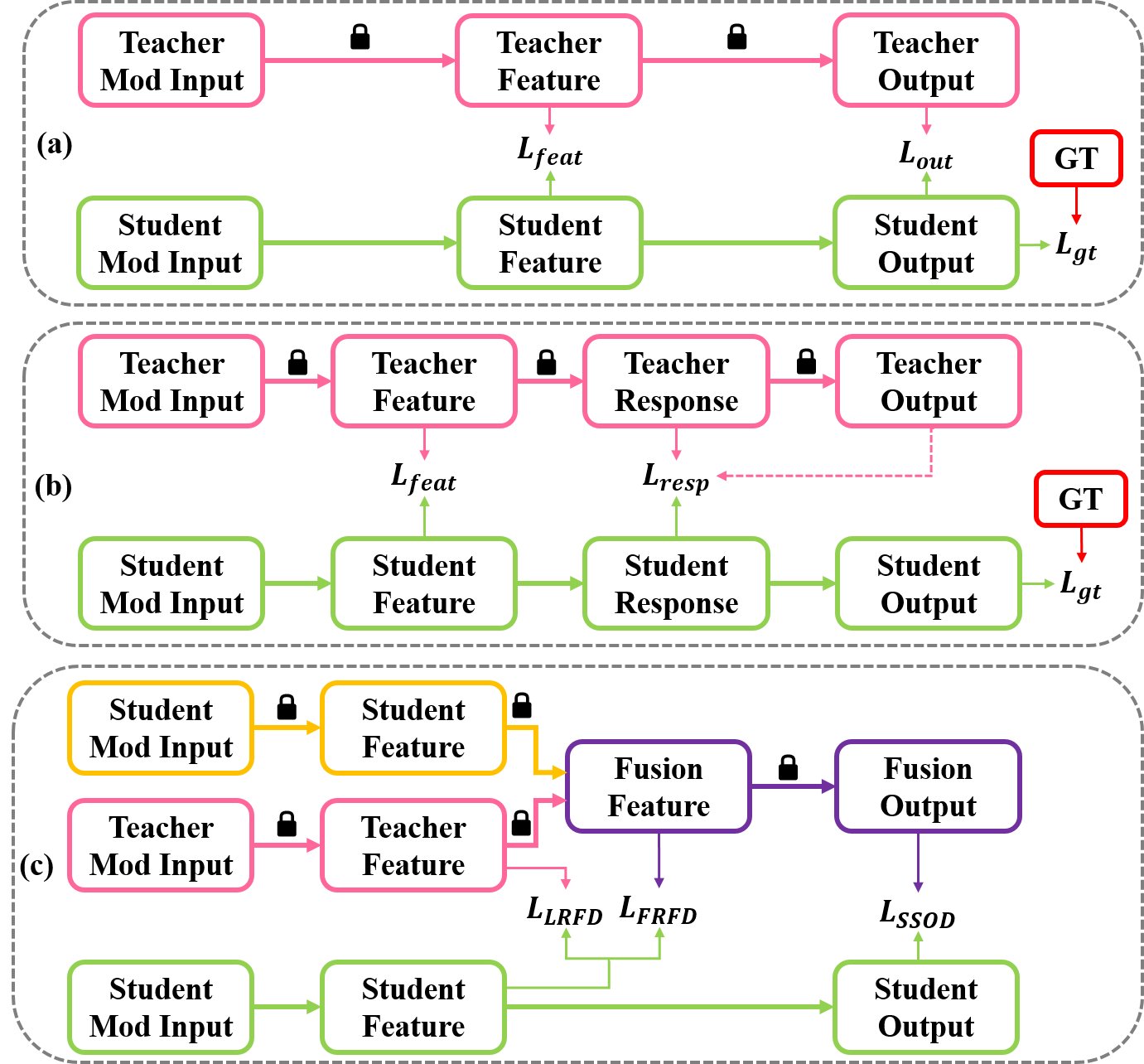}}
\caption{Comparison of the current mainstream cross-modality knowledge distillation approaches (a): BEVDistill\shortcite{chen2022bevdistill},MonoDistill\shortcite{chong2022monodistill} and (b): DistillBEV\shortcite{wang2023distillbev}, UniDistill\shortcite{zhou2023unidistill}, RadarDistill\shortcite{bang2024radardistill} with our SCKD (c).}
\label{fig1}
\end{figure}

Despite promising prospects, point clouds of 4D radar still suffer from the issues of high sparsity and flickering noise. The point density of existing 4D radar is only less than one-tenth of that of Lidar, and the ``ghost point" caused by the multi-path effect also largely degrades its range measurements. Existing approaches based on pure radar\cite{xu2021rpfa,yan2023mvfan,zheng2023rcfusion,liu2023smurf} typically employ classical backbones that are originally designed for Lidar. Without special consideration of the nature of sensors, the performance of these approaches still leaves much to be desired. Multi-modality fusion based methods\cite{nabati2021centerfusion,kim2023craft,kim2023crn,xiong2023lxl,lin2024rcbevdet,wang2022interfusion} usually behave better, but the introduction of more modalities such as camera and Lidar increases the cost of the system and decreases real-time performance.

In this paper we propose SCKD, a Semi-supervised Cross-modality Knowledge Distillation method to aggregate the merits of multi-modal fusion and real-time of radar-only based methods. The main differences between our method and the existing mainstream cross-modality distillation frameworks\cite{chen2022bevdistill,chong2022monodistill,bang2024radardistill,wang2023distillbev,zhou2023unidistill} are shown in Figure~\ref{fig1}. As shown in Figure~\ref{fig1}(a) and (b), most of the existing cross-modality distillation methods equip the teacher with another modality different from the student’s, emphasizing the idea of the knowledge transfer from a strong-input teacher to a weak-input student. However, the different characteristics of the input and the de facto feature gap between the teacher and the student are largely ignored, leading to low effectiveness in knowledge distillation. Meanwhile, they all keep the ground truth as a necessary supervision for the student. In contrast, as shown in Figure~\ref{fig1}(c), our method employs a multi-modality fusion based teacher which contains the same modality as the student. Besides improving the performance of the teacher, this manner also narrows the differences of the feature spaces, making the knowledge transfer easier. Moreover, our method no longer needs the ground truth supervision for the student, which converts the distillation into a semi-supervision manner and opens the potential for utilizing large quantities of unlabeled data.

Specifically, we design an adaptive fusion module in the teacher network to effectively fuse the feature of the Lidar and radar. Two different ways of feature distillation, namely, Lidar to Radar Feature Distillation(LRFD) and Fusion to Radar Feature Distillation(FRFD), are then proposed. Together with the Semi-Supervised Output Distillation(SSOD), the pipeline effectively fulfills the knowledge transfer between the teacher and the student. Extensive experimental results show that our SCKD outperforms the state-of-the-art methods, especially when large unlabeled data are available. 

In summary, our main contributions are as follows:

\begin{itemize}
\item We propose a novel semi-supervised cross-modality distillation framework for radar-based 3D object detection. Learning the knowledge from the teacher, simple student network can boost its performance while maintaining the real-time efficiency;
\item A Lidar and radar bi-modality teacher network embedded with adaptive fusion module is proposed to boost the performance of the teacher and reduce the difficulty of knowledge transfer;
\item LRFD and FRFD module are designed to facilitate and enhance the feature distillation;
\item Semi-supervised output distillation is proposed, which improves the performance and flexibility of the method;
\item Extensive experiments on the VoD and ZJUODset datasets are carried out for evaluation. The results show that our radar-only student network is able to boost the performance of the baseline method by a large margin and outperforms the state-of-the-art methods.
\end{itemize}

\section{Related Work}

\subsection{Lidar-based 3D object detection}
Given 3D Lidar point clouds, Lidar-based 3D object detection can roughly be divided into point-based, pillar or voxel-based, and multi-view based methods. Approaches like PointNet\cite{qi2017pointnet}, PointRCNN\cite{shi2019pointrcnn}, and 3D-SSD\cite{yang20203dssd} directly extract feature from point clouds, while methods such as VoxelNet\cite{zhou2018voxelnet}, Second\cite{yan2018second}, PointPillars\cite{lang2019pointpillars}, and Centerpoint\cite{yin2021center} divide irregular point clouds into pillars or voxels for feature extraction. The multi-view based methods fuse the features from different representations to achieve better performance.
 To mitigate the shortage of semantic information in Lidar point clouds, Lidar-image fusion based methods, e.g., PointPainting\cite{vora2020pointpainting}, PointAugmenting\cite{wang2021pointaugmenting}, MSFDFusion\cite{jiao2023msmdfusion}, and LogoNet\cite{li2023logonet} fuse the spatial and semantic features at different scales, resulting in improved 3D object detection performance.

\begin{figure*}[t]
\centerline{\includegraphics[width=1.0\linewidth]{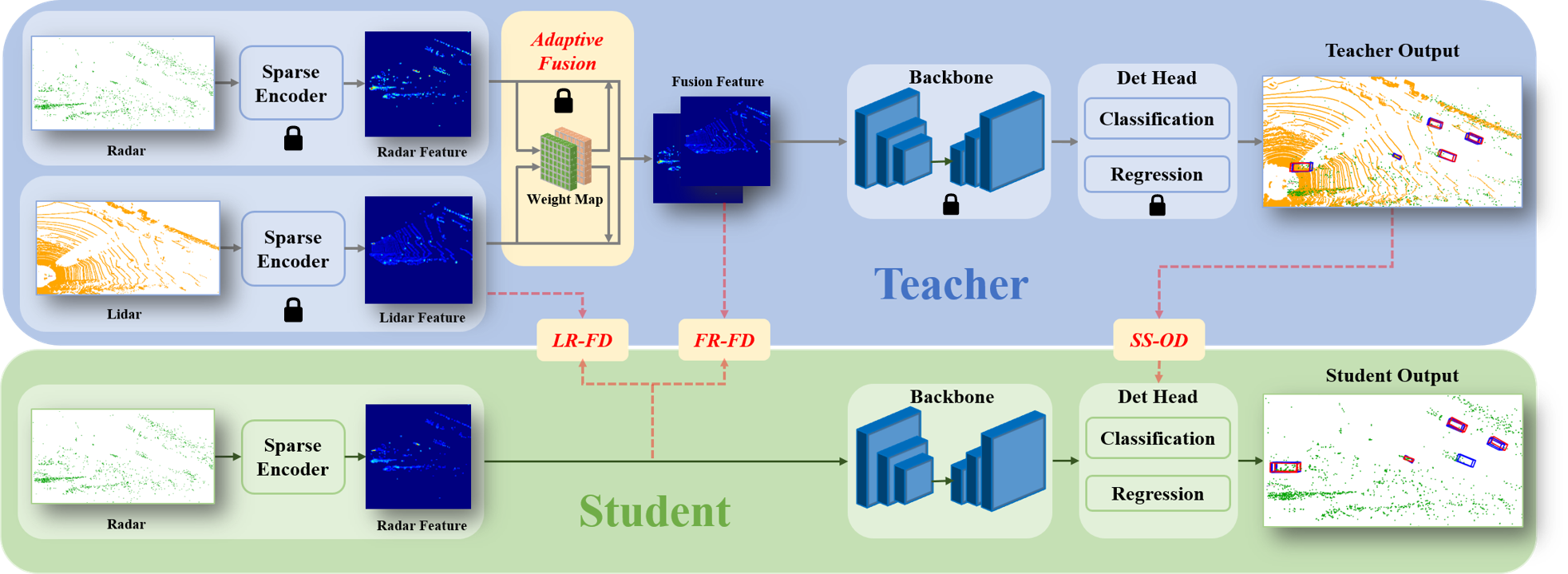}}
\caption{Overview of our SCKD Framework. The solid and dashed lines represent the data flow and calculation of the distillation loss respectively. In the inference stage, only the student network is involved.}

\label{fig2}
\end{figure*}

\subsection{Radar-based 3D object detection}
Due to the absence of height information, 3D object detection is seldom carried out on traditional 3D radar. Most of the radar-only based detection works take the 4D radar point clouds as input. RPFA-Net\cite{xu2021rpfa} utilizes a self-attention mechanism to enhance radar feature extraction. MVFAN\cite{yan2023mvfan} exploits valuable information of RCS and Doppler velocity and constructs a multi-view feature assisted detection network. Based on PointPillars, RadarPillarNet\cite{zheng2023rcfusion} designs modules specifically tailored to radar characteristics and improves the performance. SMURF\cite{liu2023smurf} introduces a novel network branch for kernel density estimation to better fuse radar feature at different levels. Due to the high sparsity of radar point clouds, more studies have focused on fusing radar with other sensors. CenterFusion\cite{nabati2021centerfusion}, CRAFT\cite{kim2023craft}, and CRN\cite{kim2023crn} primarily focus on exploring the fusion of 3D radar and images. Recently, many studies have begun to investigate the fusion of 4D radar with other sensors. RCFusion\cite{zheng2023rcfusion}, LXL\cite{xiong2023lxl}, and RCBEVDet\cite{lin2024rcbevdet} probe the fusion of 4D radar and images, while InterFusion\cite{wang2022interfusion} and ${M^2}$-Fusion\cite{wang2022multi} have ventured into the fusion of 4D radar and Lidar. Although these fusion-based methods can obtain better detection performance, they still suffer from higher sensor and computing cost, resulting in a much lower real-time performance than the radar-only methods. 

\subsection{Knowledge Distillation for object detection}

Knowledge distillation is popularly known as a model compression method, which is first applied in image classification and 2D object detection tasks\cite{yang2022masked,chen2022improved,yang2022focal,zheng2022localization,chen2022dearkd}. Recently, many knowledge distillation works have been proposed for 3D object detection. Depending on whether the same sensor modality is employed for the teacher and the student, they can be divided into common distillation\cite{zheng2021se,du2020associate,yang2022towards} and cross-modality distillation \cite{chong2022monodistill,chen2022bevdistill,wang2023distillbev,zhou2023unidistill,bang2024radardistill} methods. In the first category, SE-SSD\cite{zheng2021se} and Associate-3D\cite{du2020associate} design different data augmentation methods for the teacher network to increase the diversity of training samples. SparseKD\cite{yang2022towards} explores the impact of distillation location on the accuracy and training time of the student and finds a trade-off among them. As a cross-modality distillation method, MonoDistill\cite{chong2022monodistill} projects Lidar point clouds into images and performs feature and output distillation in the front-view. BEVDistill\cite{chen2022bevdistill} and DistillBEV\cite{wang2023distillbev} project image feature into BEV and train the image-based student with feature and instance distillation from a Lidar-based teacher. Building upon BEV, UniDistill\cite{zhou2023unidistill} proposes a universal cross-modality knowledge distillation framework, which transfers the knowledge at feature, relation and response levels, to improve the performance of camera-based or Lidar-based student detectors. The works above all need the ground truth for distillation. RadarDistill\cite{bang2024radardistill} is the very few radar-only 3D object detection network trained with a Lidar-based distiller, which has limited performance due to the lack of height information of the 3D radar. Moreover, similar to most of the existing works, it is fully supervised, which requires expensive annotated data for training. 

\section{SCKD: Semi-Supervised Cross-modality Knowledge Distillation}
In this chapter, we elaborate our proposed SCKD framework in detail. As shown in Figure~\ref{fig2}, the teacher is a Lidar-Radar bi-modality fusion network, while the student is a radar-only network. By the effective knowledge distillation of the teacher, the student can learn to extract sophisticated feature from the radar input and boost its detection performance.  We first introduce the design of the teacher network. Then, the distillation methods along with the loss function of the network are described.

\subsection{Teacher Network}

The teacher shares similar backbone with the SECOND\cite{yan2018second}, except for the bi-modality input and an adaptive fusion module. The structure of the teacher is shown in Figure~\ref{fig2}. Compared to RadarDistill\cite{bang2024radardistill}, our teacher network integrates features from two modalities, containing much richer semantic information that is helpful for knowledge distillation. Lidar point clouds $P_L$ and 4D radar point clouds $P_R$ are first encoded by voxelization and sparse 3D convolution to acquire corresponding features $F_L^T$ and $F_R^T$, respectively. The process is as follows:
$$
F_L^T = Spconv3D(Voxelization(P_L)) \eqno{(1)} $$$$
F_R^T = Spconv3D(Voxelization(P_R)) \eqno{(2)}
$$

\begin{figure}[t]
\centerline{\includegraphics[width=1.0\linewidth]{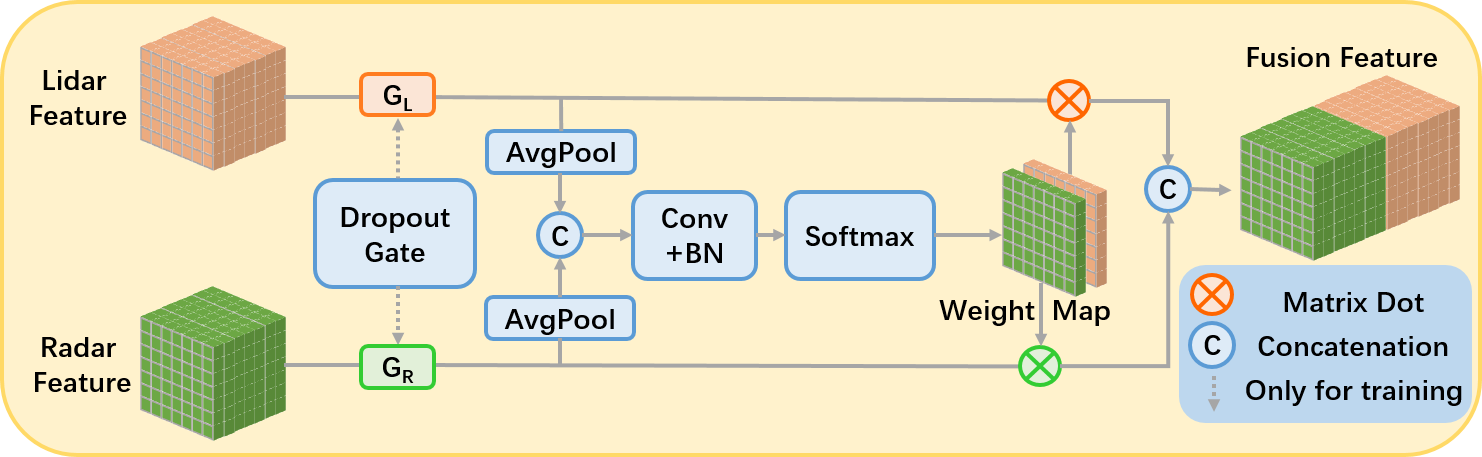}}
\caption{The structure of the Adaptive Fusion module.}
\label{figad}
\end{figure}
Considering the different nature of the Lidar and radar points, it is unwise to directly concatenate them and feed them to the consequent 2D multi-scale CNN block of the SECOND backbone. To better explore the feature of the two modalities, we propose an Adaptive Fusion (AF) module to adaptively weight and aggregate the two feature maps. As shown in Figure~\ref{figad}, the AF module is composed of a dropout gate and an adaptive weighting mechanism. The dropout gate is only effective during training and responsible for strengthening the learning of radar feature, which will be explained later. Within the adaptive fusion module, the Lidar and radar features are separately fed through average pooling, followed by convolution, batch normalization and softmax operation, resulting in adaptive weights for the corresponding input features. The final output features $F_{fusion}^T$ are then obtained by:
$$
F_{fusion}^T = Concat[W_L*F_L^T, W_R*F_R^T] \eqno{(3)}
$$
where
$$
[W_L, W_R] = Softmax(BN(Conv(F_{mix})) \eqno{(4)}
$$
with
$$
F_{mix} = Concat(AvgPool(F_L^{T}), AvgPool(F_R^{T})) \eqno{(5)}
$$
In order to further enhance the feature of each modality and promote the performance of the teacher, we introduce random dropout mechanism in the adaptive fusion module. The principle behind is that randomly discarding features of one modality allows the network to learn stronger feature extraction of another modality\cite{hwang2022cramnet}. In this case, we set a small probability $P_{drop}$ for the dropout of one modality during training. If the dropout is triggered, we set another probability $P_L$ to decide whether the Lidar or radar modality should be dropped. The process is shown as:
$$
G_L = \mathds{1}(p_1 > P_{drop}  \| \ p_2 > P_L) \eqno{(6)} $$$$
G_R = \mathds{1}(p_1 > P_{drop}  \| \ p_2 \leq P_L) \eqno{(7)}
$$

where the random numbers $p_1$ and $p_2$ are generated with a uniform probability distribution, $\mathds{1}$ is the index function. Considering the dominant role of the Lidar feature in the teacher network, the general dropout probability $P_{drop}$ and dropout probability of Lidar $P_L$ are both set to 0.2 in the experiment. It means the probability of modality dropout is 0.2, and the entire dropout probability of Lidar and radar features are 0.04 and 0.16, respectively.

The parameters of the teacher network are pre-trained and remain frozen during the training process of the student network.

\subsection{Feature Distillation}

We choose SECOND as our student model. The goal of distillation is to transfer the feature extraction capability to the student as much as possible. We propose two types of distillation, termed LRFD and FRFD, to accomplish this task. 

\paragraph{LRFD: Lidar to Radar Feature Distillation}

The feature map from Lidar contains abundant object information which can provide many hints to the extraction of student’s radar feature. Given current student feature $F_R^S$ , the LRFD takes the teacher’s Lidar feature $F_L^T$ as supervision for learning. However, directly enforcing similarities between the two types of features via a loss function tends to adversely affect the overall network performance\cite{chen2022improved}. Therefore, we feed the student's radar feature through an $Adapter^L$, which is a simple convolution layer, to simulate the Lidar feature before computing the MSE loss as:
$$
L_{LRFD} = MSE(F_{R{\rightarrow}L}^S, F_L^T) \eqno{(8)}
$$
where
$$
F_{R{\rightarrow}L}^S = Adapter^L(F_R^S) \eqno{(9)}
$$

\paragraph{FRFD: Fusion to Radar Feature Distillation} 
The fusion feature in the teacher network contains weighted Lidar and Radar feature, which is more effective for the object detection task. Comparing with the LRFD, distillation from the fused feature map has two advantages. Firstly, it contains more valuable information than the pure Lidar feature for the detection task. Secondly, the learning difficulty is lower since the fused feature itself also contains radar information. One problem is that the channel number between the fused and the radar is different and has to be aligned before distillation. Existing works\cite{chen2022improved,chen2022bevdistill,wang2023distillbev, bang2024radardistill} fulfill the channel alignment through a simple convolutional upscaling operation. However, this operation will change the original fused feature and damage the effect of distillation. We propose two separate adapters to accomplish this task. As shown in Eq. (10)~(11), two convolution-layer-based adapters, i.e., $Adapter^{L'}$ and $Adapter^{R'}$ are responsible to separately map the student’s radar feature space to the teacher’s weighted Lidar and radar feature space, as: 
$$
F_{R{\rightarrow}L}^{S'} = Adapter^{L'}(F_R^S) \eqno{(10)} $$$$
F_{R{\rightarrow}R}^{S'} = Adapter^{R'}(F_R^S) \eqno{(11)}
$$
After that, we employ the MSE loss to calculate FRFD loss $L_{FRFD}$ as:
$$
L_{FRFD} = MSE(Concat[F_{R{\rightarrow}L}^{S'}, F_{R{\rightarrow}R}^{S'}], F_{fusion}^T) \eqno{(12)}
$$

\subsection{SSOD: Semi-Supervised Output Distillation}
Most of existing distillation-based detection methods, e.g., RadarDistill\cite{bang2024radardistill}, rely on ground truth labels as the main supervision. However, ground truth labels are generally obtained through manual annotation, which can be expensive. Moreover, the existence of some difficult samples such as largely occluded objects means that directly using the ground truth labels as supervision may not bring necessary benefits to the training. To mitigate these problems, we propose to use the predictions of the teacher network as supervisions for the student network. This semi-supervised training method has two advantages. Firstly, trained by the small quantity of the labeled data, the predictions (including some false positive targets) of the teacher network are likely to provide more valuable information for the student network. Secondly, we can train the student network with much more extra unlabeled data, which can possibly improve the task performance at low costs.

Specifically, we select the detection targets $D^T$ of the teacher network based on a confidence threshold. The targets with confidence above this threshold will be regarded as pseudo-labels for the student network, as:
$$
\hat{D^T} = \mathds{1}(conf(D^T) > \sigma) * D^T \eqno{(13)}
$$
where $\sigma$ is a predefined confidence threshold. 

We then employ regular Focal loss and SmoothL1 loss to supervise the classification and regression output of the student network $D^S$, as:
$$
L_{SSOD} = L_{cls}(\hat{D^T}, D^S) + L_{det}(\hat{D^T}, D^S) \eqno{(14)}
$$
\subsection{Overall Distillation Loss}
Considering that we have eliminated the supervision from the ground truth, the student network is trained entirely by the supervision of the teacher network. The overall distillation loss is as follows:
$$
L_{total} =  \alpha L_{LRFD} + \beta L_{FRFD} + L_{SSOD} \eqno{(15)}
$$
where $\alpha$ and $\beta$ are hyper-parameters to balance the losses.

\section{Experiments}
\label{Experiments}
\subsection{Dataset and evaluation Metrics}

We conduct experiments on the popular VoD and ZJUODset datasets with accessible 4D radar and Lidar data. NuScenes\cite{caesar2020nuscenes} and TJ4DRadset\cite{zheng2022tj4dradset} datasets are not chosen because the former only contains 3D radar, and the latter's LiDAR data are not available now.

\paragraph{VoD Dataset\cite{palffy2022multi}}
The VoD dataset is currently the most popular 4D radar object detection dataset which includes Lidar, Radar, and Camera data. Following the official partitioning, we divide the training and validation set into 5139 and 1296 frames, respectively. In addition to evaluating in the entire annotated area, the dataset also requires evaluation on the driving corridor, which is a narrow region that is more likely to impact driving. To keep with previous works, we employ the AP11 evaluation metrics, and set the IOU thresholds for car, pedestrian, and cyclist to 0.5, 0.25, and 0.25, respectively. 

\paragraph{ZJUODset\cite{xu2023zjuodset}}
The ZJUODset is a dataset for long-distance 3D object detection, with the farthest detection distance reaching up to 150 meters. It also contains the data of 4D radar, Lidar and camera. Within the labeled data, we allocate 2660 frames for training and the subsequent 1140 frames for validation. We also use the rest 10640 unlabeled raw frames for semi-supervised distillation. Due to the limited samples of `pedestrian', we only evaluate on the `car' and `cyclist' categories in this dataset. The metrics utilized for evaluation are AP40, with the IOU thresholds for car and cyclist set to 0.5 and 0.25, respectively.

\subsection{Implementation Details}
For the VoD dataset, the detecting range of the network is set to [0, 51.2m] on the x-axis, [-25.6m, 25.6m] on the y-axis, and [-3m, 2m] on the z-axis. The voxel size is set to 0.05m × 0.05m × 0.1m. For the ZJUODset, the detection region is defined as [0, 158.4m], [-39.6m, 39.6m] and [-5m, 3m] on the x, y and z axis, respectively. A voxel size of 0.075m × 0.075m × 0.2m is employed for both the teacher and the student network.

We implement our SCKD based on OpenPCDet\cite{openpcdet2020} and mmdetection3d\cite{mmdet3d2020} framework. For data augmentation, we use the random flipping along the x-axis and random global scaling with the scaling factor within 0.95 and 1.05. We employ AdamW optimizer for parameter update with an initial learning rate 0.001 and a weight decay factor 0.01. The learning rate is updated with a cyclical decay method, with maximum 0.01 and minimum $10^{-7}$. The retention threshold $\sigma$ for the output distillation is set at 0.1, and the hyper-parameters $\alpha$ and $\beta$ for the loss function are both set to $3*10^{-4}$. Two NVIDIA RTX 4090 GPUs are employed during the training and distillation, with the batch size set to 8.

\subsection{Main Results}
\paragraph{Results on the VoD dataset.} The experimental results on the VoD dataset are shown in Table~\ref{tab1}. In addition to radar-only methods, we also list the methods based on the fusion of Radar and Camera for reference.

Compared with the existing radar-only methods\cite{yan2018second,yan2023mvfan,zheng2023rcfusion,liu2023smurf,deng2023see}, our approach achieves state-of-the-art performance. In contrast to the baseline method SECOND, which owns the same network structure as ours but is trained with ground truth labels instead of distillation, our approach greatly improves the mAP by 10.38\% and 6.21\% in the entire annotated area and the driving corridor, respectively. In comparison with the previously top-performing radar-based method SMURF, our SCKD also achieves an increase of 1.11\% and 2.08\% in mAP over the entire annotated area and driving corridor, respectively. The qualitative results are shown in Figure~\ref{fig4}. 

The radar-image fusion based methods usually perform better than the radar-only ones. However, they have to sacrifice the real-time performance due to the large computing costs of the fusion of image feature. It is worth mentioning that apart from the currently best-performing method LXL\cite{xiong2023lxl}, our radar-only based method surpasses the rest of those radar-image fusion methods\cite{zheng2023rcfusion,lin2024rcbevdet,chen2023futr3d,liang2022bevfusion}. Compared with LXL, we have a significant advantage in real-time performance, with more than 6 times faster than LXL in inference speed.

\begin{table*}[t!]

\begin{center}
\fontsize{9pt}{11pt}\selectfont{\begin{tabular}{c c c c c c c c c c c c c c}
\hline
 \multirow{2}*{Method}&\multirow{2}*{Ref}&\multirow{2}*{Mod}&\multicolumn{4}{c}{\textbf{Entire annotated area}}&\multicolumn{1}{c}{\textbf{}} &\multicolumn{4}{c}{\textbf{In driving corridor}}\\
\cline{4-7} \cline{9-12} 
\textbf{}&\textbf{} &\textbf{}& \textbf{\textit{Car}}& \textbf{\textit{Ped}}& \textbf{\textit{Cyc}}  & \textbf{\textit{mAP}}&  \textbf{} & \textbf{\textit{Car}}& \textbf{\textit{Ped}} & \textbf{\textit{Cyc}}& \textbf{\textit{mAP}}& \textbf{\textit{FPS}}\\
\hline
FUTR3D\shortcite{chen2023futr3d}& CVPR 2023&R+C&46.01& 35.11& 65.98& 49.03& & 78.66& 43.10& 86.19& 69.32&7.3\\
BEVFusion\shortcite{liang2022bevfusion}&ICRA 2023& R+C&37.85& 40.96& 68.95& 49.25& & 70.21& 45.86& 89.48&68.52&7.1\\
RCFusion\shortcite{zheng2023rcfusion}& TIM 2023&R+C&41.70& 38.95& 68.31& 49.65& & 71.87& 47.50& 88.33&69.23&$\backslash$\\
RCBEVdet\shortcite{lin2024rcbevdet}& CVPR 2024&R+C&40.63& 38.86& 70.48& 49.99& & 72.48& 49.89& 87.01&69.80&$\backslash$\\
LXL\shortcite{xiong2023lxl}& TIV 2023&R+C&42.33& 49.48& 77.12& 56.31& & 72.18& 58.30& 88.31&72.93&6.1\\
\hline
SECOND\shortcite{yan2018second}& Sensors 2018&R&37.46& 31.93& 55.70& 41.70& & \underline{72.04}& 42.92& 81.82&65.59&39.3 \\
MVFAN\shortcite{yan2023mvfan}& ICONIP 2023&R&34.05& 27.27& 57.14& 39.42& &69.81& 38.65& 84.87& 64.38&45.1\\
RadarPillarNet\shortcite{zheng2023rcfusion}& TIM 2023&R&39.30& 35.10& 63.63& 46.01& & 71.65& 42.80& 83.14&65.86&$\backslash$\\
LXL-R\shortcite{xiong2023lxl}& TIV 2023&R&32.75& 39.65& 68.13& 46.84& & 70.26& 47.34& \textbf{87.93}& 68.51&44.7\\
SMURF\shortcite{liu2023smurf}& TIV 2023&R&\textbf{42.31}& 39.09& \textbf{71.50}& \underline{50.97}& & 71.74& \underline{50.54}& 86.87&\underline{69.72}&$\backslash$\\
SBS\shortcite{deng2023see}& ICRA 2024&R&32.20& \underline{40.42}& 68.87& 47.03& & $\backslash$ &$\backslash$ & $\backslash$& $\backslash$&$\backslash$\\
\hline
\rowcolor{Gray}
Ours&& R&\underline{41.89}& \textbf{43.51}& \underline{70.83}& \textbf{52.08}& & \textbf{77.54}& \textbf{51.06}& \underline{86.89}& \textbf{71.80}&39.3\\
\hline
\end{tabular}
}
\caption{3D detection results on the VoD dataset. `R' and `C' stand for radar and camera, respectively. The \textbf{bold} and the \underline{underlined} separately represent the best and the second-best results among the radar-based methods.}
\label{tab1}
\end{center}
\end{table*}

\begin{table*}[t!]

\begin{center}
\begin{tabular}{c c c c c c c c c c c c c c}
\hline
 \multirow{2}*{Method}&\multirow{2}*{Mod}&\multicolumn{3}{c}{\textbf{Easy AP40@0.5/0.25}}&\multicolumn{1}{c}{\textbf{}} &\multicolumn{3}{c}{\textbf{Moderate AP40@0.5/0.25}}&\multicolumn{1}{c}{\textbf{}} &\multicolumn{3}{c}{\textbf{Hard AP40@0.5/0.25}}\\
\cline{3-5} \cline{7-9} \cline{11-13}
\textbf{} &\textbf{}& \textbf{\textit{Car}}& \textbf{\textit{Cyclist}}  & \textbf{\textit{mAP}}&  \textbf{} & \textbf{\textit{Car}} & \textbf{\textit{Cyclist}}& \textbf{\textit{mAP}}&
\textbf{} & \textbf{\textit{Car}} & \textbf{\textit{Cyclist}}& \textbf{\textit{mAP}}\\
\hline
PointPillars\shortcite{lang2019pointpillars}&R&47.99& 28.30&38.14 & & 31.24& 12.89&22.06& &16.96 &8.30 & 12.63\\
SECOND\shortcite{yan2018second}&R&56.51& 36.08& 46.30& & 35.14& 17.98&26.56& &19.15 &11.93 & 15.54\\
SCKD&R&58.50& 38.62& 48.56& &36.70& 18.57& 27.64 & &19.76 &12.63 & 16.20\\
\rowcolor{Gray}
SCKD+&R&\textbf{66.70}& \textbf{39.07}& \textbf{52.88}& &\textbf{41.72}& \textbf{21.64}& \textbf{31.68} & &\textbf{22.83} &\textbf{14.29} & \textbf{18.56}\\
\hline

\end{tabular}
\caption{3D detection results on the ZJUODset. Easy, Moderate and Hard levels evaluate the detection performance within 30 meters, 50 meters and 80 meters, respectively. `+' means we train the our student model on more unlabeled data.}
\label{tab3}
\end{center}
\end{table*}

\begin{table*}[t!]

\begin{center}
\begin{tabular}{c c c c c c c c c c c c c c c c c}
\hline
 &\multicolumn{5}{c}{Method}&\multicolumn{1}{c}{\textbf{}}&\multicolumn{4}{c}{\textbf{Entire annotated area}}&\multicolumn{1}{c}{\textbf{}} &\multicolumn{4}{c}{\textbf{In driving corridor}}\\
\cline{2-6} \cline{8-11} \cline{13-16} 
\textbf{}&\textbf{GT} &\textbf{SSOD}&\textbf{FRFD*}&\textbf{FRFD}&\textbf{LRFD}& &\textbf{\textit{Car}}& \textbf{\textit{Ped}}& \textbf{\textit{Cyc}}  & \textbf{\textit{mAP}}&  \textbf{} & \textbf{\textit{Car}}& \textbf{\textit{Ped}} & \textbf{\textit{Cyc}}& \textbf{\textit{mAP}}\\
\hline
(a)&\ding{52}&&&&&&37.46& 31.93& 55.70& 41.70& & 72.04& 42.92& 81.82&65.59 \\
(b)&&\ding{52}&&&&&40.48& 34.52& 56.10& 43.70& &72.32& 45.04& 83.28& 66.88\\
(c)&\ding{52}&\ding{52}&&&&&40.57& 35.75& 61.15& 45.82& &72.24& 46.02& 85.21& 67.82\\
(d)&\ding{52}&\ding{52}&\ding{52}&&&&41.28& 38.05& 66.55& 48.63& & 72.09& 46.89& 85.51&68.16\\
(e)&&\ding{52}&\ding{52}&&&&41.29& 39.88& 69.87& 50.35& & 72.46& 47.66& 85.66& 68.59\\

(f)&&\ding{52}&&\ding{52}&&&\textbf{41.92}& 39.91& 70.11& 50.65& & 72.53& 49.86& 85.61&69.33\\
(g)&&\ding{52}&&&\ding{52}&&41.61& 42.80& 69.79& 51.40& & 72.51& 50.38& 85.76&69.55\\
(h)&\ding{52}&\ding{52}&&\ding{52}&\ding{52}&&41.47& 40.09& 69.88& 50.48& & 72.46& 48.96& 85.53& 68.98\\
\rowcolor{Gray}
(i)&&\ding{52}&&\ding{52}&\ding{52}&&41.89& \textbf{43.51}& \textbf{70.83}& \textbf{52.08}& & \textbf{77.54}& \textbf{51.06}& \textbf{86.89}& \textbf{71.80}\\
\hline
\end{tabular}
\caption{The ablation results on the VoD dataset. GT and SSOD represent training the student with ground truth and semi-supervised output distillation, respectively. LRFD, FRFD* and FRFD denote the Lidar to radar feature distillation, fusion to radar distillation with one adapter and two adapters, respectively.}
\label{tab2}
\end{center}
\end{table*}

\paragraph{Results on the ZJUODset.} We also conduct experiments on the ZJUODSet dataset. As shown in Table~\ref{tab3}, our SCKD outperforms its competitors at all difficulty levels. Since our model is semi-supervised, we further train our model with 4 times more unlabeled raw data provided in the dataset. As expected, the performance can further be improved by up to 4.04\% mAP on the moderate level, which shows the great potential of our semi-supervised distillation mechanism.

\subsection{Ablation Study}

The ablation results are carried out on the VoD dataset, and the results are shown in Table~\ref{tab2} and Table~\ref{tab5}. 

\paragraph{Effects of SSOD.} Comparing (b) with (a) in Table~\ref{tab2}, it can be seen that the SSOD is more effective than GT in supervising the student model. Integrating both GT and SSOD can further improve the performance, as shown in (c). However, when other distillation techniques such as FRFD are introduced, leaving only the SSOD as supervision becomes more effective than using both the SSOD and GT. As shown in (e) and (d), removing GT from the student supervision leads to 1.72\% and 0.43\% mAP improvement in the entire area and the driving corridor region, respectively. More results in Table~\ref{tab5} show that our SCKD+ trained with full unlabeled data can drastically boost the performance by about 20\% mAP, which demonstrates the great benefits brought by the SSOD. 

\paragraph{Effects of LRFD and FRFD.} Comparing (f) with (e) in Table~\ref{tab2}, it can be seen that using two separate adapters in the FRFD is better than using only one. Introducing LRFD can further improve the performance, as shown in (g). Finally, integrating all SSOD, FRFD and LRFD modules can achieve the best results, with a total improvement of 10.38\% and 6.21\% mAP over the common baseline (a). 

\begin{table*}[t]

\begin{center}
\begin{tabular}{c c c c c c c c c c c c c}
\hline
 \multirow{2}*{Method}&\multirow{2}*{Teacher Trainset}&\multirow{2}*{Student Trainset}&\multicolumn{4}{c}{\textbf{Entire annotated area}}&\multicolumn{1}{c}{\textbf{}} &\multicolumn{4}{c}{\textbf{In driving corridor}}\\
\cline{4-7} \cline{9-12} 
\textbf{}&\textbf{} &\textbf{}& \textbf{\textit{Car}}& \textbf{\textit{Ped}}& \textbf{\textit{Cyc}}  & \textbf{\textit{mAP}}&  \textbf{} & \textbf{\textit{Car}}& \textbf{\textit{Ped}} & \textbf{\textit{Cyc}}& \textbf{\textit{mAP}}\\
\hline
SECOND& $\backslash$&1/4 labeled&10.70& 19.98& 11.73& 14.14& & 31.49& 29.10& 25.69&28.76 \\
SCKD& 1/4 labeled&1/4 unlabeled&10.20& 19.15& 12.60& 13.98& &28.56& 30.61& 24.47& 27.88\\
\rowcolor{Gray}
SCKD+& 1/4 labeled&Full unlabeled&\textbf{32.17}& \textbf{21.23}& \textbf{47.73}& \textbf{33.71}& &\textbf{69.73}& \textbf{31.98}& \textbf{76.28}& \textbf{59.33}\\
\hline
\end{tabular}
\caption{Performance of models trained by different amount of data on the VoD dataset.}
\label{tab5}
\end{center}
\end{table*}

\begin{table*}[t]

\begin{center}
\begin{tabular}{c c c c c c c c c c c c c}
\hline
\multirow{2}*{Teacher Modality}&\multirow{2}*{AF}&\multirow{2}*{RD}&\multirow{2}*{Student Modality}&\multicolumn{4}{c}{\textbf{Entire annotated area}}&\multicolumn{1}{c}{\textbf{}} &\multicolumn{4}{c}{\textbf{In driving corridor}}\\
\cline{5-8} \cline{10-13} 
\textbf{} &\textbf{}&\textbf{}&\textbf{}& \textbf{\textit{Car}}& \textbf{\textit{Ped}}& \textbf{\textit{Cyc}}  & \textbf{\textit{mAP}}&  \textbf{} & \textbf{\textit{Car}}& \textbf{\textit{Ped}} & \textbf{\textit{Cyc}}& \textbf{\textit{mAP}}\\
\hline
L&$\backslash$&$\backslash$&R&40.06& 40.48& 64.19& 48.24& & 72.33& 50.38& 83.67&68.79 \\
L+R&&&R&40.27& 39.01&66.63 & 48.64& & 72.35&49.80 & 85.08&69.08 \\
L+R&\ding{52}&&R&41.38& 39.56& 67.12& 49.35& & 72.44& 50.06& 85.97&69.49 \\
L+R&&\ding{52}&R&41.29& 40.31& 70.66& 50.75& & 72.53& 50.40& 86.72&69.88 \\
\rowcolor{Gray}
L+R&\ding{52}&\ding{52}&R&\textbf{41.89}& \textbf{43.51}& \textbf{70.83}& \textbf{52.08}& &\textbf{77.54}& \textbf{51.06}& \textbf{86.89}& \textbf{71.80}\\
\hline
\end{tabular}
\caption{3D detection results of different teacher network on the VoD dataset. `L' means Lidar and `R' means radar. `AF' and `RD' refer to Adaptive Fusion and Random Dropout in the teacher network, respectively.}
\label{tab4}
\end{center}

\end{table*}

\subsection{More Discussion}

\paragraph{The role of the bi-modality-fusion based teacher.} The experimental results of SCKD distilled from different configurations of teacher are shown in Table~\ref{tab4}. Compared to the Lidar-only teacher, the final Lidar-Radar-fusion teacher can improve mAP of student network by 3.84\% and 3.01\% in the entire area and driving corridor, respectively. It should thank to the richer semantic feature and narrower feature gaps of our bi-modality teacher than that of single-modality teacher. Ablation experiments on the adaptive fusion module and the random dropout modules also validate the effectiveness of our design.

\begin{figure}[h]
\centerline{\includegraphics[width=1.0\linewidth]{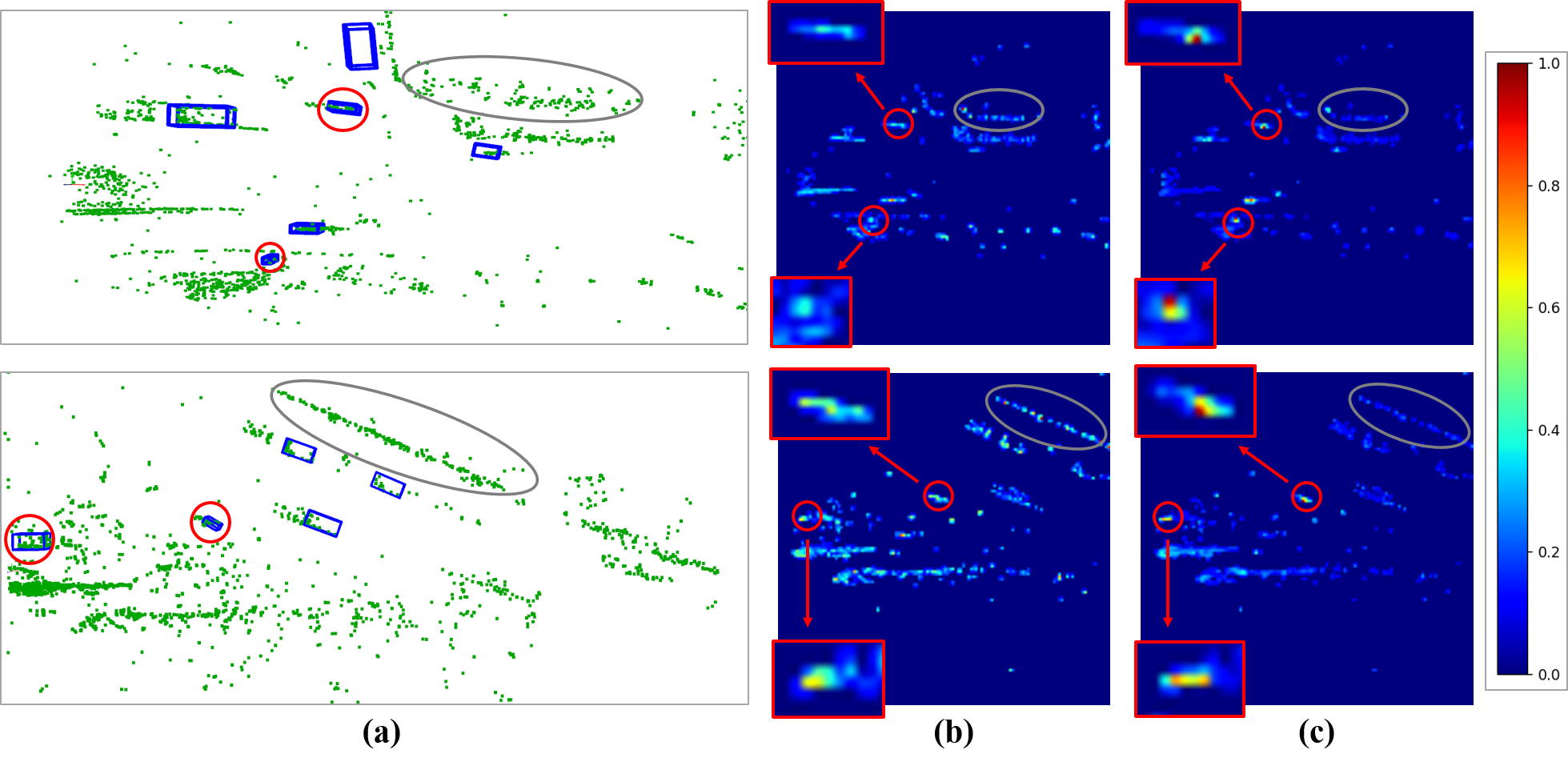}}
\caption{Comparison of the feature maps. Column (a) shows the input radar point clouds of two scenes annotated with ground truth, while (b) and (c) show the corresponding heatmaps obtained by SECOND and our method, respectively. The grey and red ellipses separately mark out the backgrounds and foregrounds for comparison.}

\label{fig3}
\end{figure}

\begin{figure}[h]
\centerline{\includegraphics[width=1.0\linewidth]{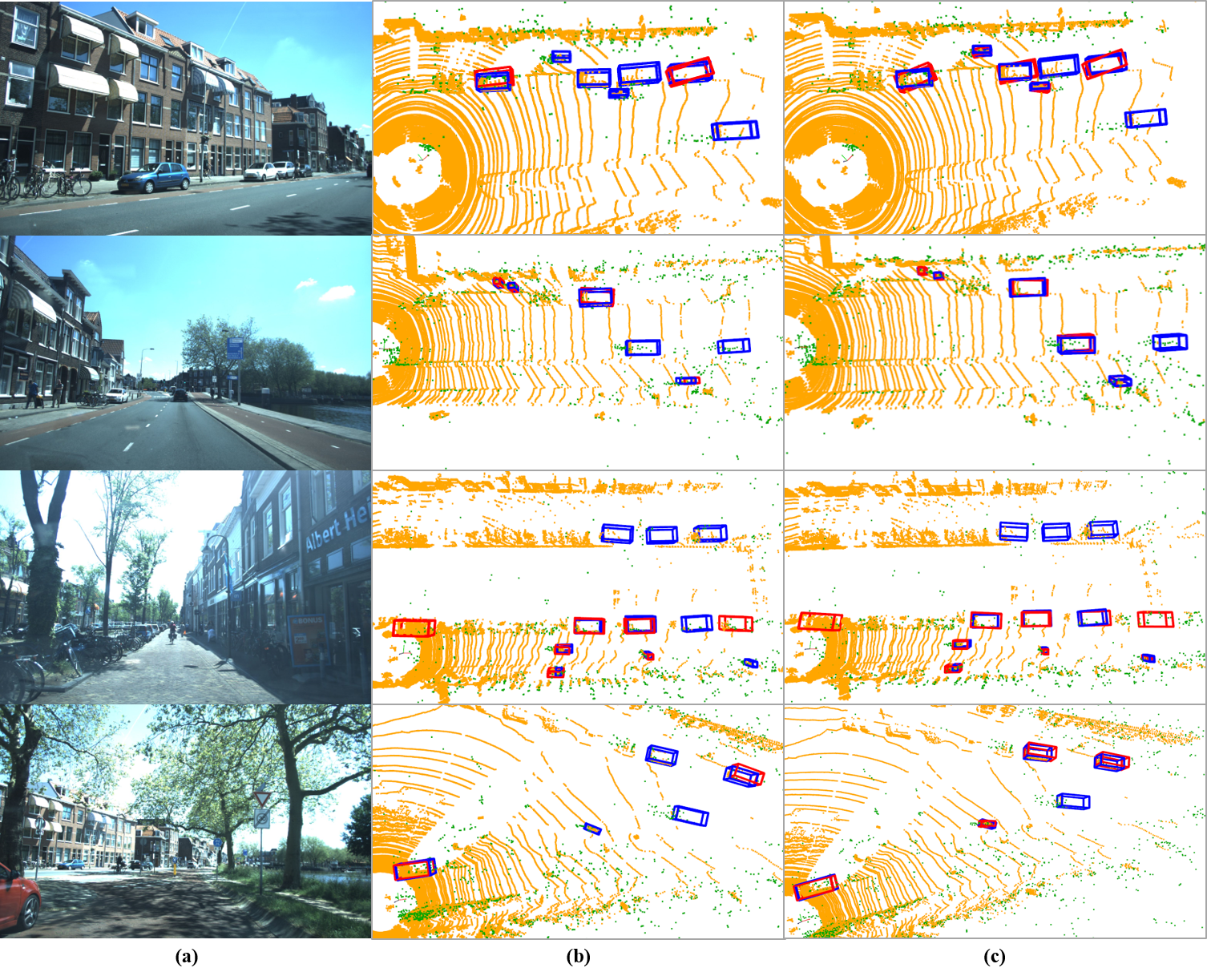}}
\caption{Qualitative results on the VoD dataset. (a) shows the scene image, while (b) and (c) show the detection results of radar-based SECOND and our method SCKD, respectively. Lidar and radar points are marked with orange and green, while the predicted and ground truth bounding boxes are colored with red and blue, respectively.}
\label{fig4}
\end{figure}

\paragraph{The visual comparison of the feature maps before and after distillation} As shown in Figure~\ref{fig3}, compared to the baseline method that only uses ground truth for training, our distillation method can drastically increase the contrast of the background and foreground in the feature heatmaps and give more attention to the foreground objects, which is vital for the final performance improvement.

\section{Conclusion}

In this paper, we propose a semi-supervised cross-modality distillation method for 3D object detection based on 4D radar-only. We design a bi-modality fusion based teacher network, which is strengthened with adaptive fusion and random dropout upon the backbone. We then design three distillation components, namely LRFD, FRFD, and SSOD, at the feature and output levels of distillation. Without introducing any computational overhead in the inference phase, our distilled student model significantly improves the object detection performance over the baseline method, surpassing all state-of-the-art radar-based methods and even most of the radar-camera fused methods. Experimental results on both the VoD and ZJUODset datasets demonstrate the effectiveness of our method.

\section{Acknowledgments}
This work was supported by The Key Research \& Development Plan of Zhejiang Province under Grant No.2024C01017, 2024C01010.

\bibliography{SCKD}

\end{document}